\documentclass[10pt,twocolumn,letterpaper]{article}

\usepackage[pagenumbers]{cvpr} 

\usepackage{graphicx}
\usepackage{amsmath}
\usepackage{amssymb}
\usepackage{booktabs}
\usepackage{xcolor,colortbl}
\usepackage{multirow}

\usepackage{hhline}
\usepackage{array}
\usepackage{graphicx}
\usepackage{multirow}
\usepackage{ragged2e}
\usepackage{booktabs}
\usepackage{hhline}

\usepackage[pagebackref,breaklinks,colorlinks]{hyperref}
\usepackage{multirow}
\usepackage{siunitx} 
\newcommand{\maxf}[1]{{\cellcolor[gray]{0.8}} #1}
\usepackage{numprint} 
\usepackage[capitalize]{cleveref}
\crefname{section}{Sec.}{Secs.}
\Crefname{section}{Section}{Sections}
\Crefname{table}{Table}{Tables}
\crefname{table}{Tab.}{Tabs.}

\def\cvprPaperID{1751} 
\def\confName{CVPR}
\def\confYear{2023}

\begin{document}

\title{JAWS: Just A Wild Shot for Cinematic Transfer in Neural Radiance Fields}
\author{Xi Wang$^1$\thanks{Equal contribution. Corresponding to jinglei.shi@nankai.edu.cn.}, Robin Courant$^{1,2}$\footnotemark[1], Jinglei Shi$^{3}$, Eric Marchand$^{1}$, Marc Christie$^{1}$\\
$^{1}$Inria, IRISA, CNRS, Univ. Rennes, $^{2}$LIX, Ecole Polytechnique, IP Paris,
$^{3}$VCIP, CS, Nankai Univ.\\
{\tt\small xi.wang@inria.fr, robin.courant@polytechnique.edu, jinglei.shi@nankai.edu.cn}\\
{\tt\small \{eric.marchand, marc.christie\}@irisa.fr}
}

\maketitle

\begin{abstract}
This paper presents JAWS, an optimization-driven approach that achieves the robust transfer of visual cinematic features from a reference in-the-wild video clip to a newly generated clip. To this end, we rely on an implicit-neural-representation (INR) in a way to compute a clip that shares the same cinematic features as the reference clip. We propose a general formulation of a camera optimization problem in an INR that computes extrinsic and intrinsic camera parameters as well as timing. By leveraging the differentiability of neural representations, we can back-propagate our designed cinematic losses measured on proxy estimators through a NeRF network to the proposed cinematic parameters directly. We also introduce specific enhancements such as guidance maps to improve the overall quality and efficiency. Results display the capacity of our system to replicate well known camera sequences from movies, adapting the framing, camera parameters and timing of the generated video clip to maximize the similarity with the reference clip. 
\end{abstract}

\section{Introduction}
\label{sec:introduction}
\begin{figure}[t]
  \centering
  \centerline{\includegraphics[width=\columnwidth]{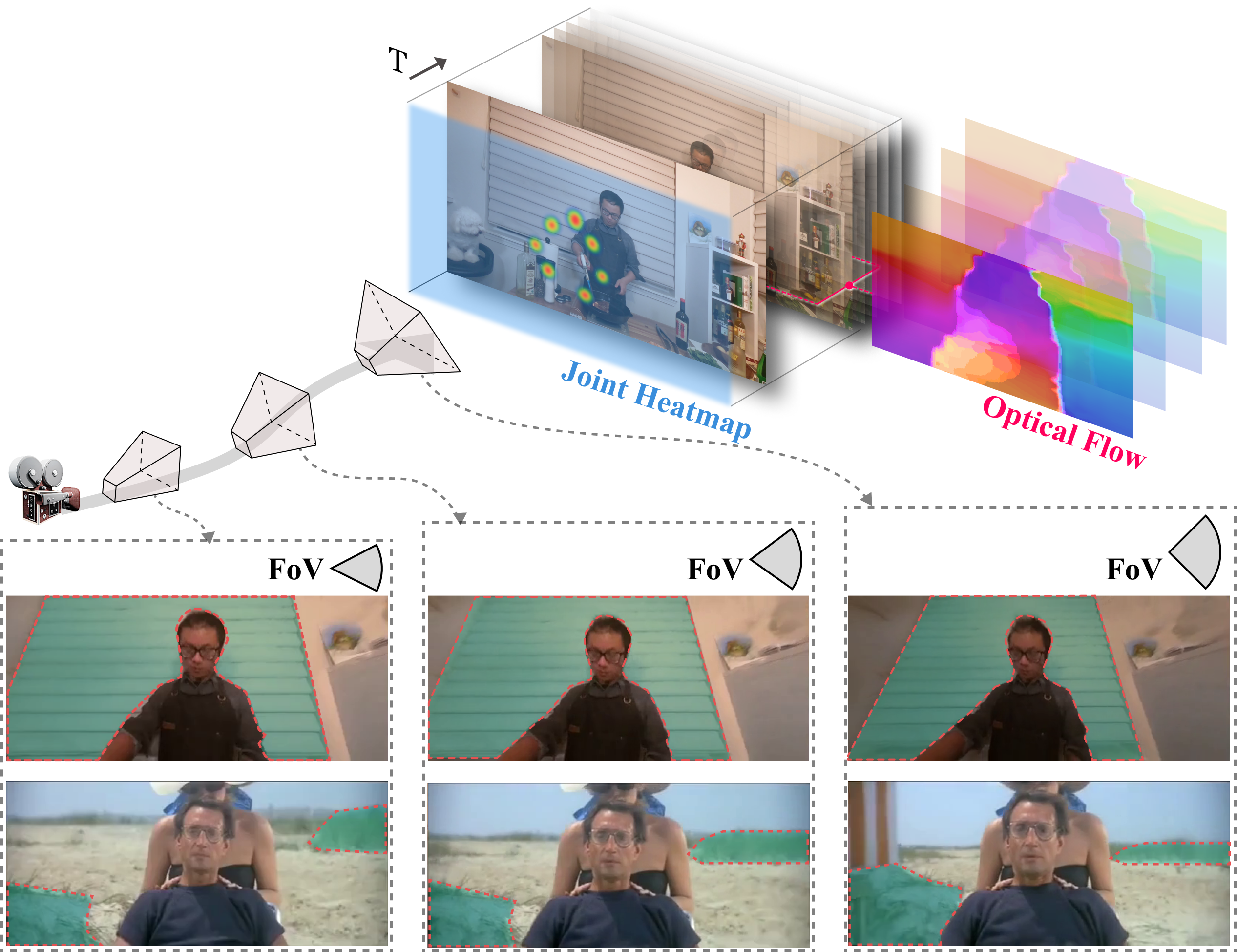}}
  \caption{This illustration displays the capacity of our proposed method to transfer the cinematic motions from the in-the-wild famous film clip \textit{Jaws} (bottom) to another context, replicating and adapting the dolly-zoom effect (a combined translation plus field of view change in opposite directions, causing distinctive motions on background and foreground contents).}
  \label{fig:teaser}
  \vspace{-0.3cm}
\end{figure}

Almost all film directors and visual artists follow the paradigm of \textit{watch-and-learn} by drawing inspiration from others visual works. Imitating masterpieces (also known as visual homage) in their visual composition, camera motion or character action is a popular way to pay tribute to and honor the source work which influenced them. This subtly draws a faint, yet distinctive clue through the development of the whole film history. Examples are commonplace: across the dizzy effect of dolly-zoom in \textit{Vertigo} to the scary scene in \textit{Jaws} (see Fig.~\ref{fig:teaser}); or from the old school Bruce Lee's kung-fu films to blockbusters such as \textit{Kill Bill} and \textit{Matrix} series. The cinematic knowledge encompassed in reference sequences is therefore carried out and inherited through these visual homages, and have even been adapted to more modern visual media such as digital animation or video games. Audiences obviously acknowledge the strong references between the epic Western \textit{The good, the bad and the ugly} of 1966 and the 2010 \textit{Red Dead Redemption}. 

The creation of such \emph{visual homages} in real or virtual environments yet remains a challenging endeavor that requires more than just replicating a camera angle, motion or visual compositions. Such a \emph{cinematic transfer} task is successful only if it can achieve a similar visual perception (\textit{look-and-feel}) between the reference and homage clip, for example in terms of visual composition (\ie how visual elements are framed on screen), perceived motion (\ie how elements and scene move through the whole sequence), but also, camera focus, image depth, actions occurring or scene lighting.

Inspired by the aforementioned \textit{watch-and-learn} paradigm, we propose to address the cinematic transfer problem by focusing on motion characteristics, \ie solving a \emph{cinematic motion transfer} problem.
 
While different techniques are available to extract camera poses and trajectories from reference videos (\eg sparse or dense localization and mapping techniques) and potentially transfer them, the naive replication of such trajectories to new 3D environments (mesh-based or implicit NeRF-based) generally fail to reproduce the perceived motion due to scaling issues, different screen compositions, lack of visual anchors, or scene dissimilarities. Furthermore, these extractiontechniques are very sensitive to widespread camera effects such as shallow depths of field or motion blur. 

In this paper, we propose to address this cinematic motion transfer problem following a different path. Rather than extracting visual features from the reference clip (as geometric properties or encoded in a latent representation) and designing a technique to then recompute a new video clip using these visual features, we rely on the differentiable nature of NeRF representations. We propose the design of a fully differentiable pipeline which takes as input a reference clip, an existing NeRF representation of a scene, and optimizes a sequence of camera parameters (pose and focal length) in both space and time inside the NeRF so as to minimize differences between the motion features of the references views and the clip created from the optimized parameters. By exploiting an end-to-end differentiable pipeline, our process directly backpropagates the changes to spatial and temporal cinematic parameters. The key to successful cinematic motion transfer is then found in the design of relevant motion features and means to improve guidance in the optimization. Our work relies on the combination of an optical flow estimator, to ensure the transfer of camera directions of motions, and a character pose estimator to ensure the anchoring of motions around a target. A dedicated guidance map is then created to draw the attention of the framework on key aspects of the extracted features. 

\noindent The contributions of our work are:

\noindent \textbf{The first feature-driven cinematic motion transfer technique} that can reapply motion characteristics of an in-the-wild reference clip to a NeRF representation.

\noindent \textbf{The design of an end-to-end differentiable pipeline} to directly optimize spatial and temporal cinematic parameters from a reference clip, by exploiting differentiability of neural rendering and proxy networks.

\noindent \textbf{The proposal of robust cinematic losses combined with guidance maps} that ensure the effective transfer of both on-screen motions and character framing to keep the cinematic visual similarity of the reference and generated clip.

\section{Related work}
\label{sec:related_work}
\noindent \textbf{Neural scene representation.}
Representing a scene efficiently and properly according to different applications, has long remained an open question in the fields of computer vision and graphics. Fueled by the development of deep learning in the past decade, typical scene representation methods such as mesh \cite{zhang2020path,nimier2020radiative}, voxel \cite{lombardi2019neural,jiang2020sdfdiff}, point cloud \cite{li2020end} or light field \cite{shi2020learning,mildenhall2019llff} showed great progress in rendering photorealistic contents. More recently, the NeRF method~\cite{mildenhall2020nerf} has drawn a huge attention from both industry and academia fields. NeRF proposes to learn attributes of light rays from multiple images, thereby encoding scene information in an implicit way using a deep neural network. Subsequent works~\cite{pumarola2021d,xian2021space,tretschk2021non,li2021neural} extended the NeRF-based methods to consider dynamic scenarios, where both temporal and spatial information are encoded by the network, enabling to synthesize images from any viewpoint and at any time.

In addition to the extension to dynamic scenes, many efforts have been made to improve network performance by either accelerating its training \& rendering processes~\cite{garbin2021fastnerf,reiser2021kilonerf,yu2021plenoctrees,muller2022instant}, deterring aliasing~\cite{barron2021mip}, extending bounds~\cite{barron2022mip}, reducing required samples~\cite{chen2021mvsnerf,roessle2022dense}. All contribute to make NeRF-based methods a powerful scene representation approach for cinematic applications. It is noteworthy that NeRF scene representations can also serve as a differential environment to inversely optimize camera poses, as with iNeRF~\cite{yen2021inerf}, where authors retrieve the reference images $SE(3)$ pose in a given NeRF scene description. Others~\cite{zhu2022nice, iMap} propose to extend the iNeRF model by addressing the localization and mapping problems.
By then adding a functional network as a proxy, the whole workflow is capable of tackling more complex computer vision tasks~\cite{mildenhall2019llff,guo2018learning,Gao_2021_CVPR}. NeRF networks can likewise work as a functional scene descriptor~\cite{Mazur:etal:ARXIV2022,li2022nerf}, followed by a predefined proxy network, to produce desired intermediate features for the final task.

\noindent \textbf{Camera control and virtual cinematography.}
\emph{Camera control} is a well established task in robotics that consists in exploiting sensor-acquired information to guide camera motion through an environment under some desired constraints, with an optimization framework. This topic covers a wide range of problems,\eg visual servoing~\cite{marchand2002controlling}, viewpoint computation~\cite{Triggs1995AutomaticCP}, or target tracking~\cite{papanikolopoulos1993visual}. 

These seminal contributions have been largely extended to account for different visual quality metrics and constraints. Problems have also been transposed from real-environments (robotics) to virtual
ones (computer graphics) by addressing virtual camera control problems~\cite{christie2008camera}. 

While most contributions encoded motion properties (speed, jerk, optical flow) as constraints on the degrees of freedom of the camera (or of the camera trajectory path~\cite{huang2016trip}), an increasing number of techniques have been exploiting real data to constrain/guide the camera motion. In~\cite{kurz2010camera}, authors propose to perform camera motion style transfer by first extracting a camera trajectory from a given film clip using structure-from-motion techniques, performing a multi-frequential analysis of the motion, and regenerating the motion from its frequential parameters in a new 3D environment. This however only \emph{replayed} the motion without adapting it the the target scene contents.

Later, through advances in deep learning techniques, contributions have started to train camera motion strategies from datasets of camera motions. Drone cinematography is a good example. 
Techniques such as Imitation Learning (IL) were used, combining the idea of Reinforcement Learning but training a model to learn from expert examples. In~\cite{loquercio2021learning}, the authors designed an IL system with simulated sensor noise to train drones to fly across in-the-wild agnostic scenarios. 

Huang \etal~\cite{Huang_2019_CVPR} exploit optical flow and human poses to guide drone cinematography controls via an IL framework. Targeting similar drone cinematic objectives,~\cite{bonatti2020autonomous} designed a DQN to control four directional actions for achieving tasks such as: obstacles avoidance, target tracking and shooting style application. Recently, Jiang \etal~\cite{jiang20sig} trained a Toric-based feature extractor from human poses in synthetic and real film data for achieving cinematic style control via a pre-trained latent space when the 3D animation is known. Following the similar topic~\cite{jiang21siga} tackled the keyframing problem which allows users to constrain the generated trajectory by a dedicated LSTM framework.

While these approaches do achieve some form of transfer of camera and motion characteristics, these are either dedicated to specific sub-fields (drone cinematography~\cite{bonatti2020autonomous,Huang_2019_CVPR}), or focus on image composition \cite{jiang20sig,jiang21siga} without considering the image motion characteristics.

\section{Preliminary knowledge}
\label{sec:preliminarly}
\noindent \textbf{Neural rendering.} Our work strongly relies on NeRF representations of 3D scenes~\cite{mildenhall2020nerf} which describe a scene through a Multiple Layer Perceptron (MLP) network by inputting different 3D locations and view directions, and inferring color and volume density. Volumetric rendering techniques then enable the reconstruction of an image from any viewpoint by retrieving the color and density of each pixel and integrating on the line direction of its pixel-correspondent rays in the MLP. A NeRF model $\mathcal{N}_\mathbf{\Theta}$ can therefore be viewed as a mapping between a camera pose $\mathbf{T}$ and a reconstructed image $\mathbf{\hat{I}}$, where $\hat{\mathbf{I}}=\mathcal{N}(\mathbf{T})$ by abusing the term and regrouping all the rays into image pixels through a camera projective model: $\mathcal{N}_{\mathbf{\Theta}}: SE(3) \rightarrow \mathbb{R}^{H \times W \times 3} ; \mathbf{T}  \mapsto  \mathcal{N}(\mathbf{T})$.

In the training, parameters $\mathbf{\Theta}$ are updated by minimizing an on-screen image loss (eg. photometric loss) on each ray, which compares pixel-level information of the reconstructed image $\mathcal{N}(\mathbf{T})$ and its associated reference view $\mathbf{I}^*$: 
\begin{equation}
    \hat{\mathbf{\Theta}} = \arg \min_{\mathbf{\Theta}} \mathcal{L}(\mathbf{\Theta} \mid \mathcal{N}(\mathbf{T}), \mathbf{I}^*)
\label{eq: trainNeRF}
\end{equation}
Training a NeRF network can therefore be seen as optimizing MLP parameters  $\mathbf{\Theta}$ from a set of known camera poses and corresponding images. 

\noindent \textbf{Camera pose inference.}
The NeRF training process can also be applied in an inverse manner (see iNeRF~\cite{yen2021inerf}), to estimate a camera pose $\hat{\mathbf{T}} \in SE(3)$ in an already trained NeRF model, against a reference view $\mathbf{I}^{*}$. 
To ensure the estimated pose still lies in the $SE(3)$ manifold, the optimization process is performed in the canonical exponential coordinate system~\cite{ma20043dvision} based on an initial camera pose $\mathbf{T}(0)$:
\begin{equation}
\begin{array}{c}
\hat{\mathbf{T}}(\theta) = e^{[{\xi}]\theta} \ \mathbf{T}(0)\quad
\text{where} \quad e^{[{\xi}]\theta} = \begin{bmatrix}
e^{[{\omega}]\theta} & K(\theta, \omega, v) \\
0 & 1 
\end{bmatrix} \\ \\
\text{with}   \quad
 K(\theta, \omega, v) = \left\{ \begin{array}{cl}
         v\theta & \text{if} \quad \omega = 0 \\
        \frac{(I - e^{[{\omega}]\theta})[{\omega}] v+\omega\omega^Tv\theta}{\lVert \omega \rVert} & \text{otherwise}
    \end{array} \right.
\end{array}
\end{equation}

\noindent where $[{\xi}]$ is a twist matrix, $(\omega, v)$ are twist coordinates, and $\theta$ is a magnitude. The optimization problem is then changed to seek for the optimal camera parameters $(\hat{\theta}_k, \hat{\omega}_k, \hat{v}_k)$. 

\section{Method}
\label{sec:method}
\begin{figure*}
  \centering
  \includegraphics[width=\textwidth]{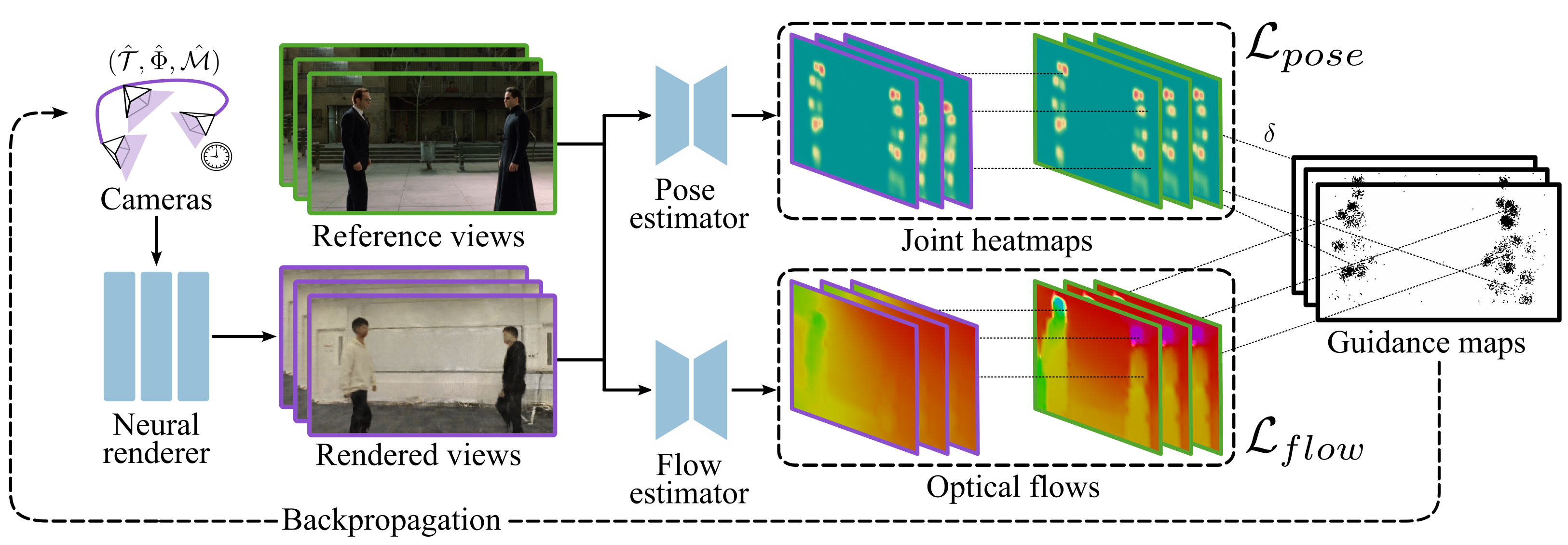}
  \caption{Overview of JAWS pipeline. Given the cinematic parameters (camera motion, focal length and timing parameters) to optimize $(\hat{\mathcal{T}}, \hat{\Phi}, \hat{\mathcal{M}})$, we first synthesize views through a fixed  NeRF, then compute joint heatmaps and optical flows from the rendered (\textit{purple}) and reference (\textit{green}) views, via pose and flow estimators respectively. A guidance map (\textit{black}) is calculated and helps the sampling of pixel with gradient s.t. cinematic parameters can be updated through a backpropagation tracing back to all proxy and NeRF networks.}
  \label{fig:main}
  \vspace{-0.3cm}
\end{figure*}

\subsection{Cinematic transfer}

Inspired by the photorealistic recording capability of neural rendering methods (\eg NeRF) and the related camera pose estimation methods such as iNeRF, in this paper we explore the possibility of proposing a NeRF-based \textit{Cinematic Motion Transfer} from in-the-wild references, following the \textit{watch-and-learn} paradigm explained in Sec.~\ref{sec:introduction}. Such a problem consists in extracting cinematic information from a reference clip and reapplying it in another scene s.t. the rendered visual content shares high cinematic similarity and effects. We propose our method, JAWS, displayed in Fig.~\ref{fig:main}, to address the \textit{Cinematic Transfer} goal in a reference agnostic NeRF environment. The main idea is to optimize multiple cameras' cinematic parameters through the design of robust cinematic losses in a differentiable framework. To improve the performance of this optimization process, we extended our work with multiple techniques. 

\label{subsubsec: formulation of the problem}
\noindent \textbf{Formulation of the problem.} Similar to the formulation of Eq.~\ref{eq: trainNeRF}, we describe our problem as an inverted optimization. However, contrasting with iNeRF~\cite{yen2021inerf} which only optimizes a single camera pose and photometric loss, we target multiple cameras' cinematic parameters using two different, yet complementary, losses to ensure the transfer of key cinematic characteristics to synthesized clip. These robust cinematic losses are: (i) an \textit{on-screen} loss $\mathcal{L}_{s}$, which guides the framing and position of significant elements; and (ii) an \textit{inter-frame} loss $\mathcal{L}_{i}$, guiding the correct overall motion along the sequence.

We therefore re-formulate the problem as an inverted optimization of 
$N$ cameras $\mathcal{T}=\{\mathbf{T_i}\}_N$ w.r.t. robust cinematic losses $\mathcal{L} = \mathcal{L}_{s}+\mathcal{L}_{i}$ between a set of synthesized views $\mathcal{N}(\mathcal{T})$ and reference film clip images $\mathcal{I}_{r}$:

\begin{equation}
\begin{array}{c}
    \hat{\mathcal{T}} = \arg \underset{\mathcal{T}}\min \: \mathcal{L}(\mathcal{N}(\mathcal{T}) \mid \mathcal{I}_{r}, \mathbf{\Theta})
\end{array}
\label{eq:target_formular}
\end{equation}

The definition of cinematic camera parameters will be discussed in Sec.~\ref{subsubsection: Cinematic parameters}, and details of our proposed cinematic losses will be given in Sec.~\ref{subsection: Cinematic_loss}. We also propose specific enhancement techniques to further improve the efficiency of our system in Sec.~\ref{framework_enhancements}.

\label{subsubsection: Cinematic parameters}
\noindent \textbf{Cinematic parameters.} Film shooting cameras are usually more complex than relying on a $SE(3)$ camera model: variable focal length is often employed to realize many creative film effects, such as \textit{whip zoom} (rapid zooming to close up on actors facial area to express surprise and draw sudden focus) or \textit{dolly-zoom} (translating the camera in compensated direction to suggest strong emotional impact). 

In addition to camera parameters, another factor we need to take into consideration is that film shooting always comprises the dynamic nature of actors and scene props, which are rarely discussed in most NeRF-based pose estimation works~\cite{yen2021inerf,avraham2022nerfels,zhu2022nice}. They only adopt static environmental hypothesis.
It therefore appears mandatory to include extra timing parameter $m$ representing the temporal condition in dynamic scenes, and focal length $\phi$ into our cinematic parameters with 6DoF camera pose, in order to address properly the cinematic motion transfer problem. Fortunately, thanks to the recent progress in extending static NeRF to the dynamic scene with the help of temporal deformation representation~\cite{pumarola2021d,xian2021space,tretschk2021non}, dynamic NeRF methods can render unseen views at a specific instant (\ie timing parameter $m$) for animated scenes or objects. One of our main intuitions behind trying to handle dynamic scenes is to rely on another MLP to encode warp information of rays such that the displacement of pixels can be retrieved from a canonical space (\ie classic static NeRF methods). With this differentiable temporal MLP, we can follow the same optimization idea to further manipulate the timing of characters dynamic motion or scenes changing in the neural rendering system.

We therefore include intrinsic parameters, especially the focal length for each camera $\Phi = \{\phi_i\}_N$, and the timing moments, \ie temporal parameters $\mathcal{M} = \{m_i\}_N$ for dynamic scenes. We then transform the Eq.~\ref{eq:target_formular} into the following by relying our cinematic camera parameters:
\begin{equation}
\begin{array}{c}
    \hat{\mathcal{T}}, \hat{\Phi}, \hat{\mathcal{M}} = \arg \underset{{\mathcal{T}, \Phi, \mathcal{M}}}\min \: \mathcal{L}(\mathcal{N}(\mathcal{T}, \Phi, \mathcal{M}) \mid \mathcal{I}_r, \mathbf{\Theta})
\end{array}
\label{eq:target_formular2}
\end{equation}

\subsection{Cinematic losses}
\label{subsection: Cinematic_loss}

Two challenges hinder the understanding and transferring of cinematic motion from reference clips to a different scene: (i) the difference of content in terms of the relative scale, mismatched appearance and even dissimilar dynamic motion of characters; (ii) the necessity to describe and compare the motions of the cameras between each other s.t. generated sequence shares high visual dynamic similarity.

As described in Sec.~\ref{subsubsec: formulation of the problem}, we propose a combination of two different, yet complementary, optimization criteria, which we term as \textit{on-screen} and \textit{inter-frame} cinematic losses $\mathcal{L} = \mathcal{L}_{s}+\mathcal{L}_{i}$.
Intuitively, they cover \textit{framing} and \textit{camera motion} aspects respectively. 

\noindent \textbf{On-screen loss.} Our on-screen loss collects and transfers the on-screen information to ensure framing consistency. To this end, we decide to anchor the framing to the one of most significant entities on screen, human characters, like many previous works~\cite{Huang_2019_CVPR,bonatti2020autonomous,jiang20sig}. Exploiting character on-screen poses benefits from: (i) its sparse and robust information in comparison to pixel-wise criterion, and (ii) being descriptive enough to match corresponding key-regions between two views from different scenes. 
By contrast with similar tasks~\cite{jiang20sig, Huang_2019_CVPR}, which only extract the final human pose keypoints, in our work we exploit the detection confidence heatmaps predicted by a deep neural network. On the contrary to singular keypoints resulting from non-differentiable operations, those heatmaps are differentiable and contain richer information.

To compute a differentiable distance between two heatmaps, we use Wasserstein distance~\cite{kantorovich1960mathematical} (w-distance for short). 
The choice of w-distance aims to highlight the on-screen spatial relation (similar to on-image Euclidean joints distance) rather than MSE-like measures emphasizing on intensity difference (see more in~\cite{arjovsky2017wasserstein}). By minimizing the w-distance between two character pose heatmaps, we are able to transfer the framing information from a target reference to a differentiable rendering space. 

Lets consider a confidence heatmap $\mathbf{H} \in (\mathbb{R^+})^{H \times W \times J}$ with $J$ being joint number on human skeleton (see Fig.~\ref{fig:main}) generated from a pose estimation network, where each channel represents a detection probability of a given joint in the image domain. To calculate the character pose loss $\mathcal{L}_{\text{pose}}$, we compute w-distance $d_w$ between confidence maps of a reference $\mathbf{H}^*$ and a synthetic view $\hat{\mathbf{H}}$ on each channel.
\begin{equation}
\begin{array}{c}
    \mathcal{L}_{\text{pose}} = \sum\limits_{i=1}^{J}d_w(\mathbf{H}^*_i, \hat{\mathbf{H}_i}) + ||S(\mathbf{H}^*)-S(\hat{\mathbf{H}})||\\[5pt]
    \text{where} \quad S(\mathbf{H}) = \sum\limits_{i=1}^J\sum\limits_{j=1}^J d_w(\mathbf{H}_i, \mathbf{H}_j)
\end{array}
\end{equation}
The regularization $S$ is defined as an inter-joint matrix measured in w-distance of each joint $\mathbf{H_j}$ to all the others from the \textit{same} heatmap $\mathbf{H}$. This term assures the inter-joint shape similarity between heatmaps while optimizing the framing.

\begin{figure}
\begin{center}
  \includegraphics[width=0.47\textwidth]{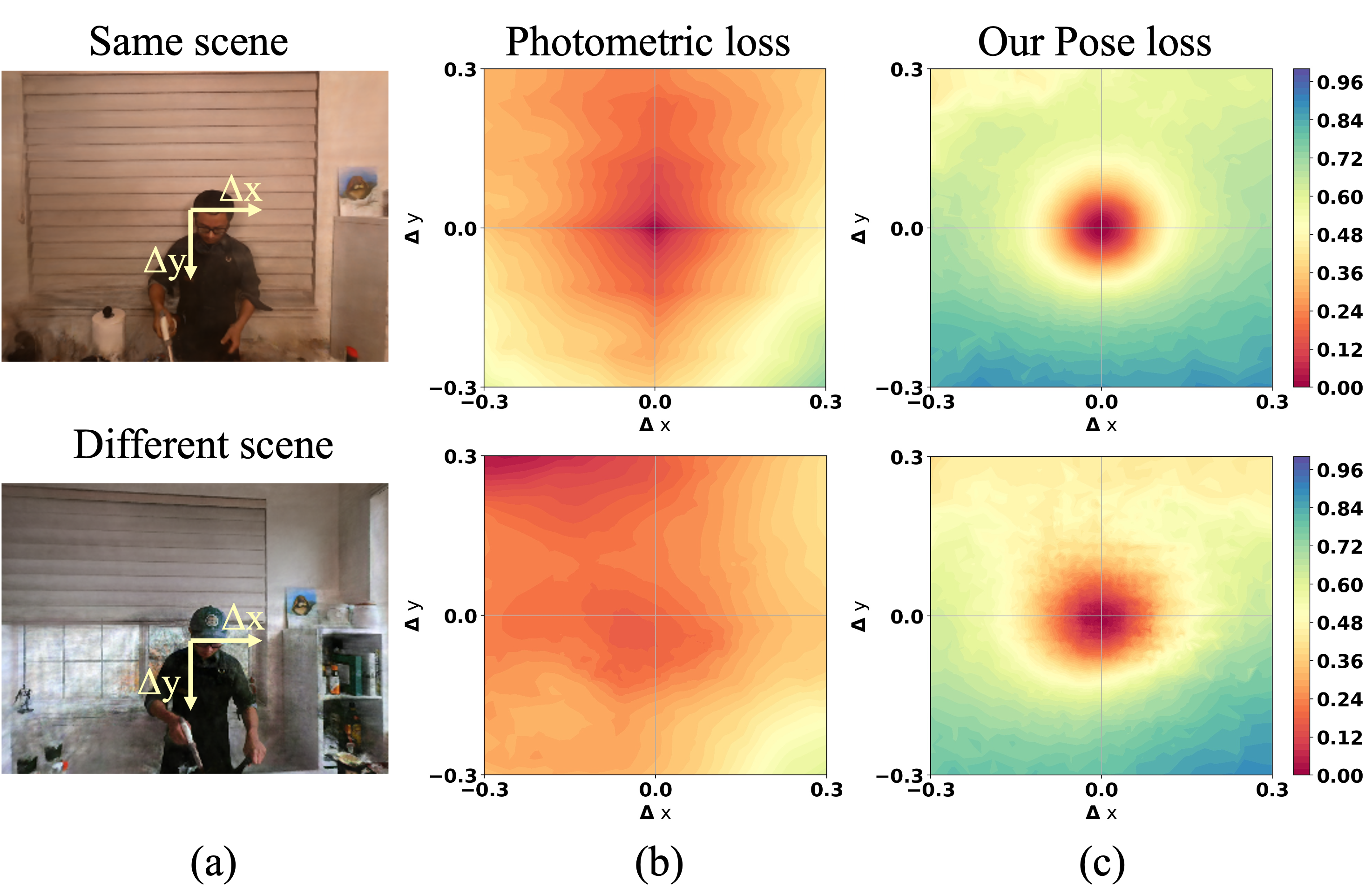}
  \caption{Normalized photometric loss \emph{vs.} our pose loss under different camera position perturbations in the horizontal and vertical directions (x and y axis in col (b,c)) in the NeRF scene. First row shows loss convergence when applying the reference image to the same scene; and the second row shows results when testing the same reference under a different scene.
  }
  \label{fig:loss distribution}
 \end{center}
 \vspace{-0.7cm}
\end{figure}

Fig.~\ref{fig:loss distribution} illustrates the loss distribution when perturbing the camera pose around a reference position on $x$ and $y$ translational directions ($\Delta X, \Delta Y$), within same or different scene against the reference. We compare our proposed character pose loss to the photometric loss, which is commonly used in inverted NeRF methods~\cite{yen2021inerf,zhu2022nice}. Results demonstrate that the convergence cone of our loss preserves invariant while measuring the reference to a different content and provides correct human-centered framing information.

\noindent \textbf{Inter-frame loss.} On the other side, we present another criterion that depicts the relative camera motion performance along the sequence dimension of synthesized views.

In order to incorporate camera motion, we propose to extract the optical flow between two consecutive frames for estimating the similarity between reference and synthesized views. It takes advantages of (i) describing the motion within frames: enabling to infer indirectly the camera trajectory; and (ii) being agnostic towards frame content: allowing to compare views from different scenes.
 
In this work, we choose a differentiable deep neural network generated optical flow (see Fig.~\ref{fig:main}) for the purpose of back-propagating to the designed cinematic parameters through the neural rendering framework. To calculate the flow loss $\mathcal{L}_{flow}$, we compute the endpoint distance~\cite{baker2011database} between the reference $\mathbf{O}^*$ and synthetic $\hat{\mathbf{O}}$ sequences of flows:

\begin{equation}
    \mathcal{L}_{\text{flow}} =  \lVert \mathbf{O}^* - \hat{\mathbf{O}}\rVert_2
\end{equation}

\noindent \textbf{Optimization loss.} The final loss for the optimization problem $\mathcal{L}_{\text{total}}$ is a linear combination of the pose and flow loss, corresponding to the \textit{on-screen} and \textit{inter-frame} loss termed in Eq.~\ref{eq:target_formular2} respectively: $\mathcal{L}_{\text{total}} = \alpha \mathcal{L}_{\text{pose}} + \beta \mathcal{L}_{\text{flow}}$.

We adjust $\alpha$ and $\beta$ during the optimization to achieve robust and dynamic emphasis on the two aspects separately.


\subsection{Framework enhancements}
\label{framework_enhancements}
\noindent \textbf{Sampling strategy.}
Using pre-trained proxy networks to achieve high-level information is popular when solving complex computer vision tasks. The differentiable nature of neural networks allows the loss to be backpropagated to the upperstream networks such that high-level constraints can influence the whole system. 
Similarly, we design our on-screen and inter-frame losses, s.t. they are proxied by human pose~\cite{wang2022lite} and optical flow~\cite{teed2020raft} estimation networks. 

However, two main factors hinder the direct connection of proxy networks to the NeRF-based methods: (i) backfire happens on the memory usage and computational complexity since the images from NeRF methods are composed in ray-pixel elements, each pixel links to all MLP parameters when back-propagating, yielding amplified memory according to pixel number;
(ii) similar to avoiding textureless regions in computer vision tasks, our proposed loss is not uniformly distributed across the whole image field (especially the character pose loss). The divergence can be triggered by non-informative image regions.

Many NeRF-based camera estimation papers~\cite{yen2021inerf,zhu2022nice} rely on sampling skills to accelerate the computation, lighten the memory and focus on relevant regions.
Unfortunately, similar experiences can not be directly transplanted to our problem, since proxy networks expect \textit{entire} input image rather than sparsely sampled pixels.

\begin{figure}
 \begin{center}
  \includegraphics[width=0.47\textwidth]{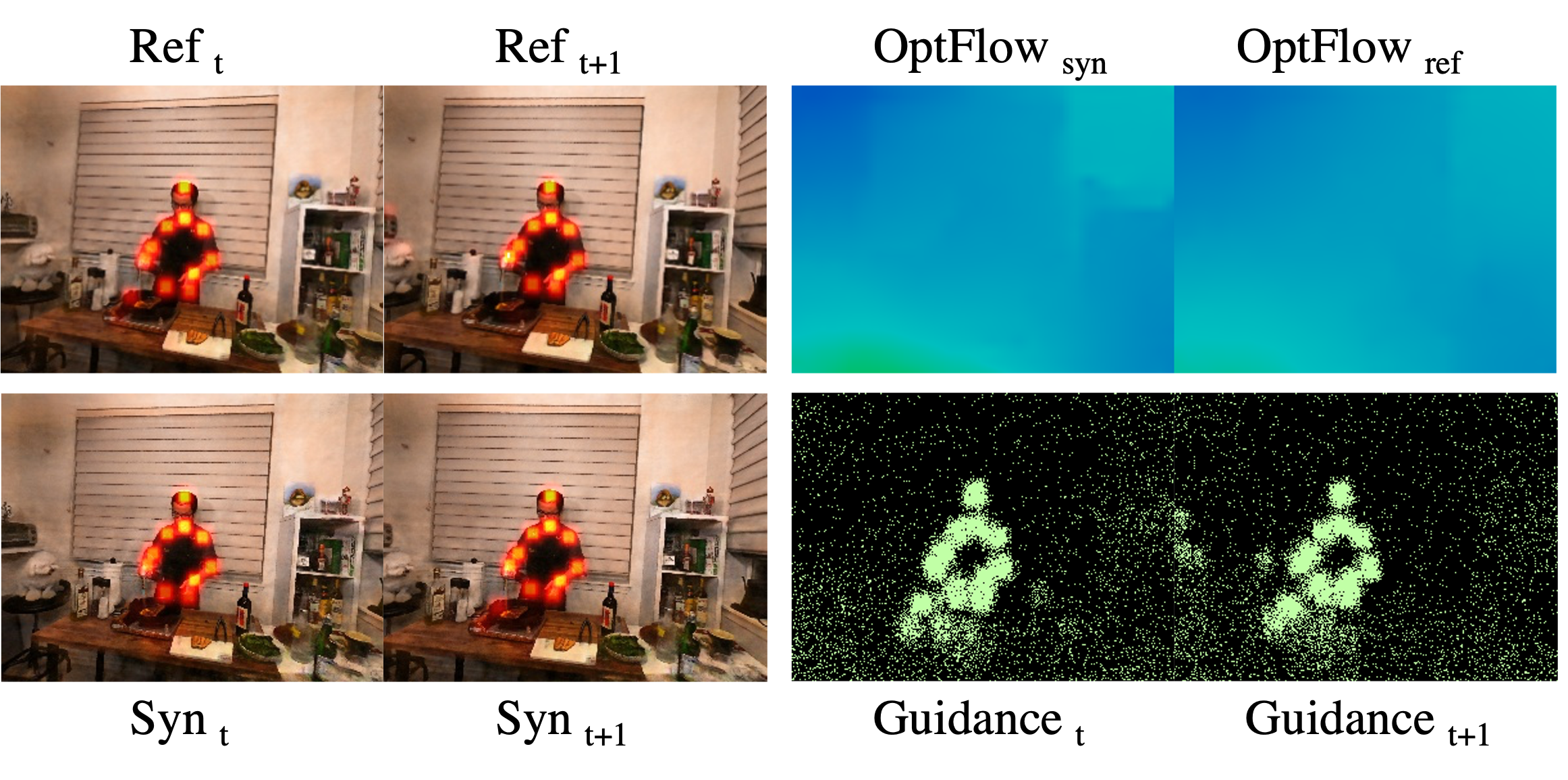}
  \caption{An example of computation of guidance maps (\textit{bottom-right}) for two cameras respectively, via uniting the difference of heatmaps (\textit{left}) and optical flows (\textit{top-right}). High intensities on the guidance maps represent the sampled pixels \textit{with} gradient.}
  \label{fig:guidance}
 \end{center}
\vspace{-0.5cm}
\end{figure}

We therefore come up with our guidance map skill to address this problem: instead of keeping only sampled pixels, we detached the gradient w.r.t. a probabilistic guidance map to achieve: (i) lighter memory usage and fast computation; (ii) attention-like mechanism to help convergence focus on informative regions. More specifically, objective parameters are updated by referring gradients backpropagated from only a Gaussian sampled small portion of pixels according to the intensity from guidance map during the optimization, whereas \textit{all} pixels contribute during the inference of proxy networks. In contrast to iNeRF~\cite{yen2021inerf} using keypoint-driven regions on color image, our guidance map $\textbf{G}$ (Fig.~\ref{fig:guidance}) is computed as united differential map between the generated human pose heatmaps and optical flows in order to highlight the key-regions contributing to the gradient:
\begin{equation}
\begin{array}{c}
    \mathbf{G} = f_{n}(|\mathbf{O}^*-\hat{\mathbf{O}}|) + f_{n}(|\mathbf{H}^*-\hat{\mathbf{H}}|)
\end{array}
\end{equation}
where $f_{n}$ is min-max normalization, and $|.|$ the absolute value.

\section{Experiments}
\label{sec:experiments}
\noindent \textbf{Implementation details.}
Our implementation is based on `torch-npg'~\cite{torch-ngp} and Pytorch~\cite{pytorch}, with Adam optimizer~\cite{kingma2014adam} and `GradNorm' algorithm~\cite{chen2018gradnorm} to adjust loss weights during the optimization. We used InstantNGP~\cite{muller2022instant} and D-NeRF~\cite{pumarola2021d} for neural rendering, RAFT~\cite{teed2020raft} for the optical flow estimation and LitePose~\cite{wang2022lite} to infer the joint heatmaps. Cameras are optimized two by two, and linked with a sliding window.
For more details see Suppl. material. 

\noindent \textbf{Datasets.} 
We test on multiple heterogeneous datasets to show the effectiveness and generality of our proposed method: including Blender-made dynamic dataset from D-NeRF~\cite{pumarola2021d}, multi-view light-field-like video dataset from~\cite{li2022neural,zhang2021stnerf}, mobile phone recorded video from nerfstudio~\cite{nerfstudio}, and our Unity synthesized animation renderings.

\subsection{Qualitative results}

\begin{figure*}
  \includegraphics[width=\textwidth]{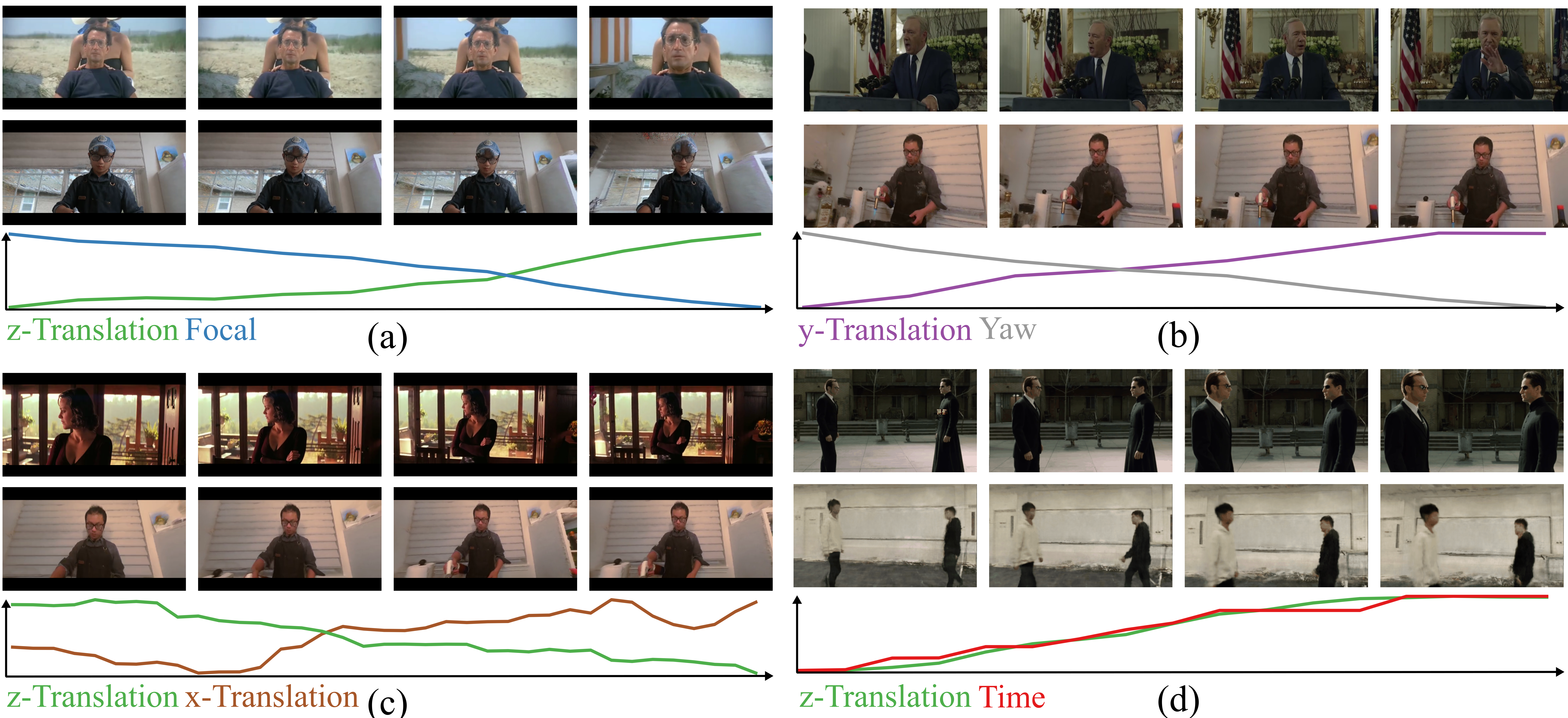}
  \caption{Examples of movies to NeRF transfers (\textit{frames}), and below the evolution along clips of: time (\textit{red}), focal length (\textit{blue}) and x-translation (\textit{brown}), y-translation (\textit{purple}), z-translation (\textit{green}) and yaw (\textit{grey}). 
  (a) a dolly-zoom in \textit{Jaws}; (b) an arc in \textit{House of Cards}; (c) handheld pull-out in \textit{Inception}; and (d) a push-in in \textit{The Matrix Reloaded}. x is along the width, y along the height, and z along depth.}
  \label{fig:gallery}
\end{figure*}

\noindent \textbf{Movies to NeRF.} Fig.~\ref{fig:gallery} shows examples of cinematic transfers from a classic shots to synthesized clips. Reference keyframes are sampled for optimizing the proposed cinematic parameters, interpolation is then applied on calculated parameters for high fps of final rendering.

\noindent In Fig.~\ref{fig:gallery}: (a) highlights the ability of our pipeline to search for the optimal focal length in addition to the camera pose: making it possible to reproduce the well-known \textit{Vertigo} effect, i.e. pushing forward and zooming out. With the parameters evolution (below \textit{frames}), we show that the focal length (\textit{blue}) and the z-translation (\textit{green}) tend to compensate each other, realizing the visual effect of ``retreating background and forwarding foreground"; (b) shows that JAWS can also reproduce complex arc motion, composed of yaw (z-rotation, \textit{grey}) and y-translation (\textit{purple}) at the same time; (c) demonstrates the aspect to mimic subtle camera motions, such as handheld movements. Evolution of x- and z-translation (\textit{brown} and \textit{green}) present the shaky effect with irregular progressions; (d) illustrates that our pipeline can align the temporal axis to fit synthesized content with the reference. The timing parameter (\textit{red}) increases with the z-translation (\textit{green}), s.t. both characters approach to each other while the camera is forwarding, to present similar framing to the reference. In addition to the shown examples, more demos are available in the Suppl. material.


\begin{figure}
  \includegraphics[width=0.47\textwidth]{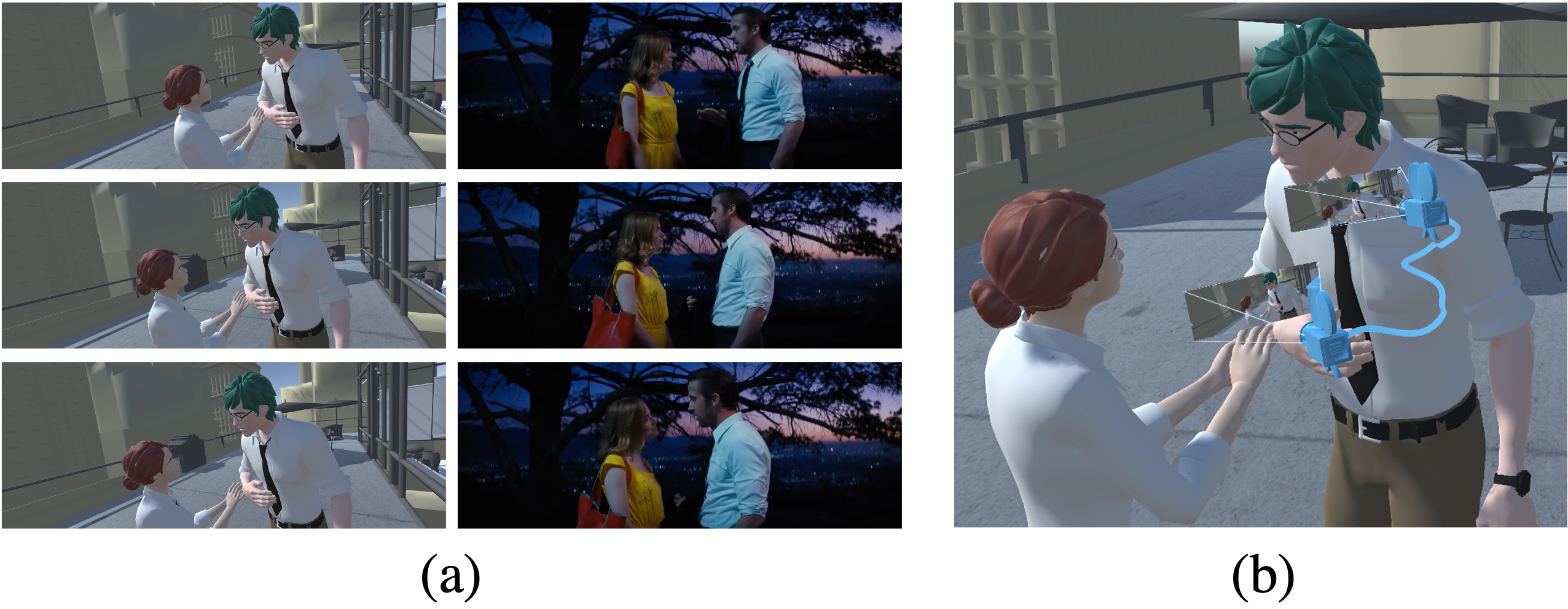}
  \caption{we show that the neural rendered results share high \textit{look-and-feel} quality to the original reference (see (a)), and the exported trajectory (b) can be easily manipulated by artists.}
  \label{fig:unity}
\end{figure}
\noindent \textbf{Movies to 3D engine.} We propose a possible application of our method on animation workflow: using JAWS to generate referenced camera trajectories to guide the animation shooting in graphics engine \eg Unity3D. It consist of: (i) training a NeRF by multi-view images rendered by Unity3D; (ii) generate desired cinematic transfer trajectory from JAWS according to a reference clip; (iii) re-apply the generated trajectory into Unity3D for rendering high quality animation. The advantage of combining two systems is to bridge the indifferentiability in graphics engine and lower animation quality in dynamic NeRF. See Fig.~\ref{fig:unity} for the rendered results presenting high visual similarity with the original reference, and the exported trajectory can be easily used by artists as a reasonable starting point of their workflow.

\subsection{Ablation study}

\begin{figure}
 \begin{center}
  \includegraphics[width=0.47\textwidth]{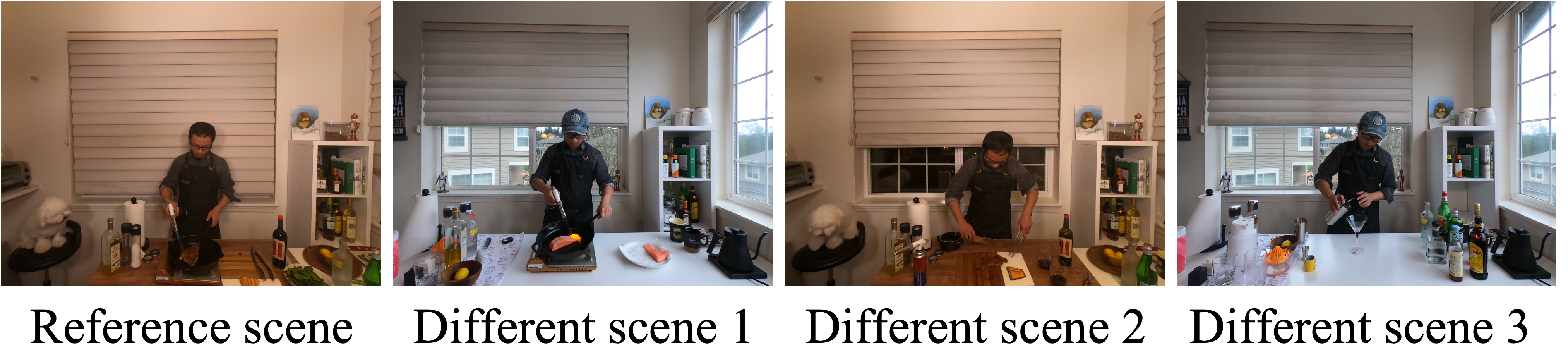}
  \caption{Scenes tested during the ablation study.}
  \label{fig:ablation_loss}
 \end{center}
 \vspace{-1cm}
\end{figure}

\begin{table*}
    \centering
    \resizebox{1.0\textwidth}{!}{
    \begin{tabular}{l|SSSSSS|SSSS|SSSS|SSSS}
    \toprule
    \multirow{3}{*}{\textbf{Loss}} & \multicolumn{6}{c|}{\textbf{Same scene}} & \multicolumn{4}{c}{\textbf{Different scene 1}} & \multicolumn{4}{c}{\textbf{Different scene 2}} & \multicolumn{4}{c}{\textbf{Different scene 3}}\\
    & \multicolumn{3}{c}{\cellcolor{blue!25}{Same init}} & \multicolumn{3}{c|}{\cellcolor{red!25}{Different init}} & \multicolumn{2}{c}{\cellcolor{blue!25}{Close init}} & \multicolumn{2}{c}{\cellcolor{red!25}{Different init}} & \multicolumn{2}{c}{\cellcolor{blue!25}{Close init}} & \multicolumn{2}{c}{\cellcolor{red!25}{Different init}} & \multicolumn{2}{c}{\cellcolor{blue!25}{Close init}} & \multicolumn{2}{c}{\cellcolor{red!25}{Different init}}  \\
    & \textit{ATE} $\downarrow$ & \textit{JE} $\downarrow$ & \textit{PE} $\downarrow$ & \textit{ATE} $\downarrow$ & \textit{JE} $\downarrow$ & \textit{PE} $\downarrow$ & \textit{ATE} $\downarrow$ & \textit{JE} $\downarrow$ & \textit{ATE} $\downarrow$ & \textit{JE} $\downarrow$ & \textit{ATE} $\downarrow$ & \textit{JE} $\downarrow$ & \textit{ATE} $\downarrow$ & \textit{JE} $\downarrow$ & \textit{ATE} $\downarrow$ & \textit{JE} $\downarrow$ & \textit{ATE} $\downarrow$ & \textit{JE} $\downarrow$ \\
    \midrule
    inerf~\cite{yen2021inerf}     & \maxf{1.70}    & \maxf{2.571}  & \maxf{0.171}   & \maxf{2.63} & 2.583 & \maxf{0.2327}   & /    & /    & / & /  & 24.973 & 19.365 & 32.575 & 145.267 & 32.365 & 23.998 & 32.035 & 47.340  \\
    pixel     & 3.00 &  2.823 & 0.189  & 7.8504 & \maxf{2.332} & 0.3116   & / & /   & / & / & 35.981 & 121.985 & 36.603 & 19.375 & / & / & 30.814 & 29.234\\
    pose      & 32.552 & 3.663   & 2.309 & 39.9396 & 5.470  & 2.3620 & 28.8156 & 17.877    & 25.299 & 7.018  & 31.671 & 16.061 & 32.548 & 19.858 & 36.304 & 21.925 & 31.829 & 20.121  \\
    flow      & 14.8332 &  3.229 & 1.0986   & 33.5388 & 17.221 & 4.3792   & 19.231    & 18.102    & 21.374    & 28.472  & \maxf{7.471} & 12.786 & 19.436 & 19.439 & 15.792 & 29.018 & 19.801 & 33.817  \\
    flow+pose & 9.143 &  2.864 & 1.143  & 14.8963 &  3.590615199866638 & 2.0520778010702796   & \maxf{14.7816} & \maxf{6.268}    & \maxf{14.0182} & \maxf{7.000}   & 13.782 & \maxf{8.452} & \maxf{19.385} & \maxf{9.906} & \maxf{15.338} & \maxf{7.623} & \maxf{14.927} & \maxf{10.265} \\
    \bottomrule
    \end{tabular}}
    \caption{Ablation of losses. We report absolute pose trajectory (\textit{ATE}) in cm (alignment~\cite{umeyama1991least} is applied for the trajectory from different scenes), average joints error (\textit{JE}) in pixel and image pixel-wised error (\textit{PE} ($10^{-2}$)). \textbf{/} means experiment fails to finish the trajectory. } 
    \label{tab:loss_ablations}
\end{table*}

\noindent \textbf{Loss ablation.} In order to demonstrate the effectiveness of each component and design of our system, we mainly test three aspects of our proposed method: 

\noindent (i) a video copying task is first undertaken, consists in transferring a NeRF rendered reference video (20 keyframes) which the trajectory is known, under the same and different scenes (see Fig.~\ref{fig:ablation_loss}), with same/close and different camera initialization positions. The objective is to demonstrate the robustness and characteristics of our two complementary cinematic losses across different scenes and influenced by different level of perturbations (\ie. initial position).

\noindent (ii) we carry out an experiment for retrieving correct timing and focal length information from images rendered with known parameters respectively. This experiment is to prove the ability of recovering these parameters by inverted optimization method on dynamic NeRF networks.

\noindent (iii) we investigate the influence of the guidance map by collecting the motion performance and memory usage of different sampling number (\ie number of pixels with gradient) to demonstrate that the guidance can help the convergence and mitigate the memory usage simultaneously.

Tab.~\ref{tab:loss_ablations} reports the ablation of losses for different setups: (i) same/different target scenes to the reference to emphasize the cross-domain transfer ability; (ii) same/different initial positions to highlight the robustness. Three metrics are computed: Absolute Trajectory Error (RMSE-ATE) reflects the quality of retrieved motion on $SE(3)$ similar to many tracking works~\cite{iMap,zhu2022nice}. To depict the on-screen composition similarity, we measure Pixel Error (PE) (for the same scene) and average Joint Error (JE) in pixel computed by Litepose (with a more accurate model).

We study several losses combinations: iNeRF~\cite{yen2021inerf}, \ie pixel loss with keypoint-driven sampling (w/o guidance map); pixel loss~\footnote{\textit{pixel loss} uses the guidance map to exploit gradients only for selected pixels but the loss is computed with \textit{all} pixels, whereas \textit{iNeRF} computes loss on selected ones}, pose and flow loss followed by combination of flow and pose losses (\textit{flow+pose} in Tab.~\ref{tab:loss_ablations}).

According to Tab.~\ref{tab:loss_ablations}, we can observe:
(i) iNeRF and pixel losses behave similarly: they both show good performance and robustness against perturbation for the same scene on motion (ATE) and composition qualities (PE, JE). However pixel-based methods fail frequently (\ie camera moving to non-defined area of the NeRF and yielding numerical error) under different scenes. For the barely succeeded experiments, the performances are low due to the misled pixel information by mismatched appearance from different scenes. (ii) Comparing to pixel-based methods, the others (\textit{pose}, \textit{flow}, \textit{flow+pose}) show invariance against appearance changing. Nevertheless, they all act differently: (a) flow loss tends to drift heavily if the initialization position is far to the correct one (JE and PE). The phenomenon is due to the fact that the flow focus on the inter-frame information and extracts no hint on the compositing; (b) in contrast, pose loss completely ignores the inter-frame motion and causes lower performance on ATE yet relatively better results in PE and JE on the same scene, suggesting possible ambiguities on similar human pose and different camera parameters. Heatmap pose feature also shows robustness against initial perturbation, with a small difference of ATE between the close and the different initializations; (c) by combining the two complementary losses: pose and flow, we achieve overall better performances than pose and flow separately, especially under different scene condition. Yet under the same scene, the performance is lower than pixel level tracking (iNeRF and pixel), reflecting on PE and ATE this is due to less sharpen convergence cone (see Fig.~\ref{fig:loss distribution}) limited resolution on extracted heatmap, and possible ambiguities.


\begin{figure}
  \includegraphics[width=0.47\textwidth]{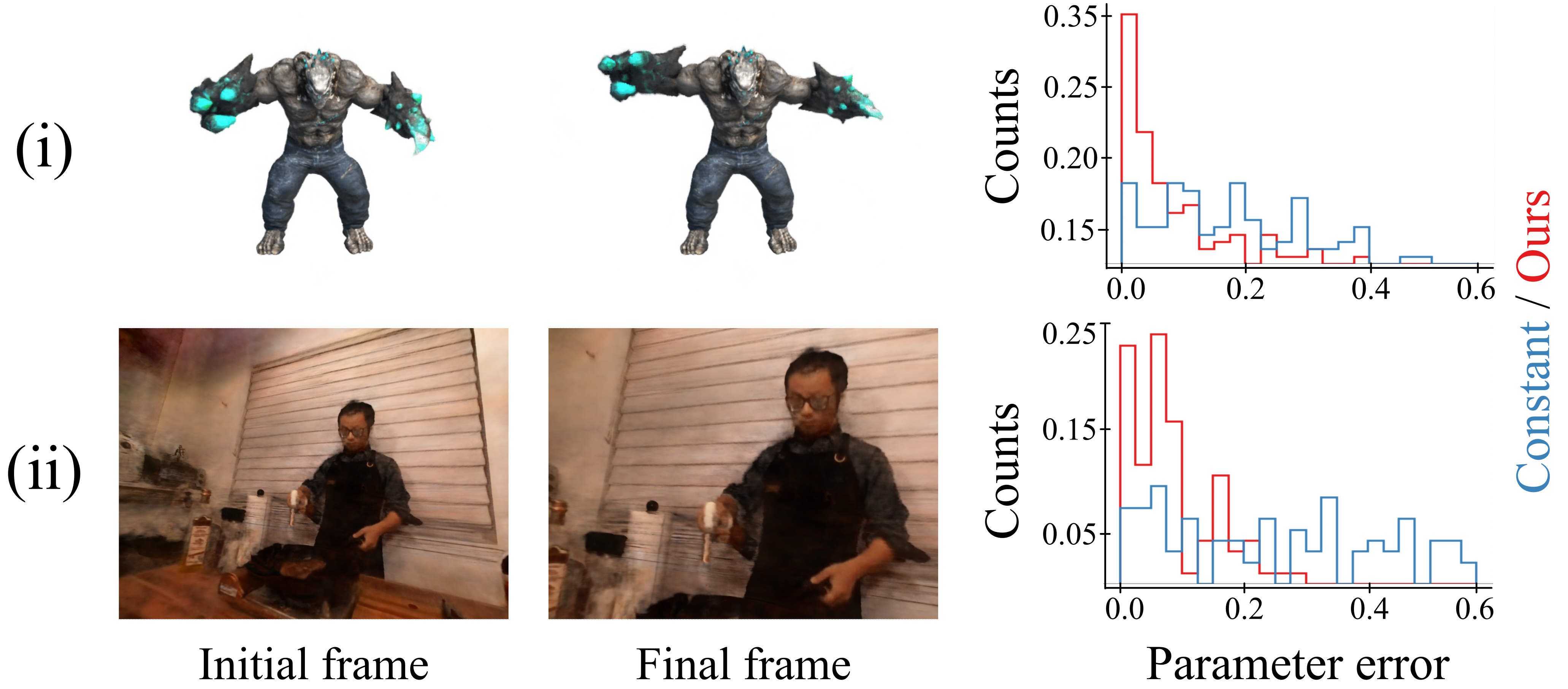}
  \caption{Visualization of (i) temporal ablation, and (ii) focal length ablation. We show the initial and final frames of the six-frames clips used for ablation. The right column shows the error distribution on the ablated parameter (time and focal length) using our method (red) and a constant parameter setup (green).}
  \label{fig:f-t-ablation}
\end{figure}

\noindent \textbf{Parameter ablation.} We also undertake ablation for parameters: time $m$ and focal length $\phi$. The experiment is done by applying our method on 6-frames clips, where only time or focal length varies. Each experiment is run 16 times with a random initial value for the studied parameters ($m$ or $\phi$). For each studied parameter, we show in Fig.~\ref{fig:f-t-ablation} the distribution of the error between the prediction and the ground-truth (\textit{red}). We compare our results with the controlled case where the studied parameter remains constant, \ie equals to the random initial value, all along the 6-frames clips (\textit{blue}).

\noindent Fig.~\ref{fig:f-t-ablation} (i) shows the result for the time ablation, where only the arms of the mutant waving and Fig.~\ref{fig:f-t-ablation} (ii) shows the result for the focal length ablation, camera pose is fixed but the focal length varies. For both examples, our prediction error (\textit{red}) is smaller than the controlled one (\textit{blue}). Some large error values for our method are due to extreme initial points, making the convergence harder. Since we are optimizing \textit{all} cinematic parameters, some effects may have ambiguities with other parameters than the studied one. 


\setlength{\tabcolsep}{2pt}
    \begin{table}[]
    \resizebox{1.0\columnwidth}{!}{%
    \begin{tabular}{ll|cccccccccc}
    \toprule
     & No. Guidance      & 500  & 1000 & 2000 & 4000 & 6000 & 8000 & 10000 & 32000 & 66752 &  \\ \cline{2-12} 
     & Avg APE (cm)      & 35.0 & 27.6 & 25.3 & 21.0 & 23.4 & 29.8 & 32.6  & 23.0  & 14.3  &  \\
     & Std APE (cm)      & 0.70 & 0.88 & 0.80 & 0.77 & 0.42 & 1.61 & 1.99  & 0.55  & 0.60  &  \\
     & Mem usage (MB) & 7273 & 7439 & 7457 & 7849 & 8339 & 8871 & 9121  & 14071 & 21151 & \\
    \bottomrule
    \end{tabular}
    }
    \caption{We show the influence of guidance map sampling number to the performance and memory usage.}
    \vspace{-0.2cm}
    \label{tab:guidance_ablations}
\end{table}
\noindent \textbf{Guidance ablation.} Tab.~\ref{tab:guidance_ablations} shows the influence of the guidance map on the performance and memory usage. The experiment is done in similar methodology to the Tab.~\ref{tab:loss_ablations} under the same scene with different initial position (see more in Suppl. Material). Low ATEs are reported when the sampling number is insufficient. An experimental optimum is around $4000$ as we used for our method. We notice higher std and memory usage when the sampling number keeps increasing. The performance re-improves when including the gradient of \textit{all} pixel in the image (66752). The low-yield behavior of higher sampling numbers could be due to the confused the gradient direction by non-informative pixels.


\section{Discussion}
\label{sec:discussion}

Despite good performance and robustness across different scenes, our method still shows limitations from: (i) highly mismatched character poses may compromise inter-frame motion and lead to undesired results;
(ii) the temporal resolution and duration of D-NeRFs are still insufficient compared to graphics engine. Generating complex and long trajectory wrt active character motion can be laborious.

\noindent \textbf{Acknowledgements.} We extend our gratitude to Anthony Mirabile and Nicolas Dufour for their invaluable contributions in constructing the 3D scene and proofreading.

{\small
\bibliographystyle{unsrt}
\bibliography{egbib,main}
}

\end{document}